\documentclass{article}
\usepackage{spconf,amsmath,graphicx}
\usepackage{times}
\usepackage{latexsym} 
\usepackage{multirow}
\usepackage{graphicx}
\usepackage{color, colortbl}

\usepackage{subcaption}

\newcounter{notecounter}
\newcommand{\enotesoff}{\long\gdef\enote##1##2{}}
\newcommand{\enoteson}{\long\gdef\enote##1##2{{
\stepcounter{notecounter}
{\large\bf
\hspace{1cm}\arabic{notecounter} $<<<$ ##1: ##2
$>>>$\hspace{1cm}}}}}
\enoteson
\enotesoff

\def\figref#1{Figure~\ref{fig:#1}}
\def\figlabel#1{\label{fig:#1}\label{p:#1}}

\def\tabref#1{Table~\ref{tab:#1}}
\def\tablabel#1{\label{tab:#1}\label{p:#1}}

\def\seclabel#1{\label{sec:#1}}
\def\eqref#1{Eq.~\ref{eqn:#1}}

\newcommand{\stackexchange}{SE}

\newcommand{\newcite}{\cite}

\usepackage{url}

\definecolor{Gray}{gray}{0.9}
\definecolor{LightCyan}{rgb}{0.88,1,1}

\title{UNSUPERVISED DOMAIN ADAPTATION OF CONTEXTUAL EMBEDDINGS FOR \\LOW-RESOURCE DUPLICATE QUESTION DETECTION}

\name
 {Alexandre Rochette*\thanks{\:\:\:\:*Equal contribution.}, Yadollah Yaghoobzadeh*\footnotemark[1], Timothy J. Hazen}
 \address {Microsoft\\ \texttt{\{alrochet, yayaghoo, TJ.Hazen\}@microsoft.com}}

\begin{document}
%\ninept
%
\maketitle
\begin{abstract}

Answering questions is a primary goal of many conversational systems or search products. While most current systems have focused on answering questions against structured databases or curated knowledge graphs, on-line community forums or frequently asked questions (FAQ) lists offer an alternative source of information for question answering systems. Automatic duplicate question detection (DQD) is the key technology need for question answering systems to utilize existing online forums like StackExchange. Existing annotations of duplicate questions in such forums are community-driven, making them sparse or even completely missing for many domains.
Therefore, it is important to transfer knowledge from related domains and tasks. Recently, contextual embedding models such as BERT have been outperforming many baselines by transferring self-supervised information to downstream tasks.
In this paper, we apply BERT to DQD and advance it by unsupervised adaptation to StackExchange domains using self-supervised learning. We show the effectiveness of this adaptation for low-resource settings, where little or no training data is available from the target domain.
Our analysis reveals that unsupervised BERT domain adaptation on even small amounts of data boosts the performance of BERT.
%We also demonstrate that unsupervised fine-tuning of BERT
%on large number of domains can further improve the  performance for a variety of target domains.
\end{abstract}

\section{Introduction}\label{intro}

Answering questions is a primary goal of many conversational systems or search products. While most current systems have focused on answering questions against structured databases or curated knowledge graphs, on-line community forums or frequently asked question (FAQ) lists offer an alternative source of information for question answering systems to utilize. For example, \emph{StackExchange (\stackexchange{})} is a popular community forum website containing posted questions and community supplied answers that span many different domains.\footnote{https://stackexchange.com/}  To take advantage of such data sets, conversational systems need the ability to recognize when a question asked by a user is semantically identical to a previous asked and answered question contained within a forum site or FAQ list.

In this paper, we address the problem of duplicate question detection (DQD) \cite{burke1997question,jeon2005finding,lei2016,nakov-semeval2016}.
There are two main application scenarios of DQD in the forums:
(i) given a database of questions, find the semantically equivalent duplicates;
(ii) given a question as a query, rank the database of questions based on their pair-wise similarity to the query.
Both cases are important for efficient information seeking.

For learning DQD models, we need question pairs annotated with duplicate labels. In \stackexchange{}, expert users do these annotations voluntarily. In practice, the annotations are sparse or even missing for many domains. It is therefore important to transfer knowledge from related tasks and domains in order to perform DQD in new domains. With deep learning models such domain adaptation is typically achieved with various forms of transfer learning (i.e., fine tuning models learned from other tasks and domains to new tasks and domains). Adversarial domain adaptation~\cite{Ganin15,arjovsky2017wasserstein} has also been successfully applied to multiple domains to improve cross-domain generalization in DQD substantially~\cite{shah2018adversarial}.

% alrochet: I'm not sure i understand the intent of the sentence below:

\begin{figure}
    \centering
    \includegraphics[width=170pt]{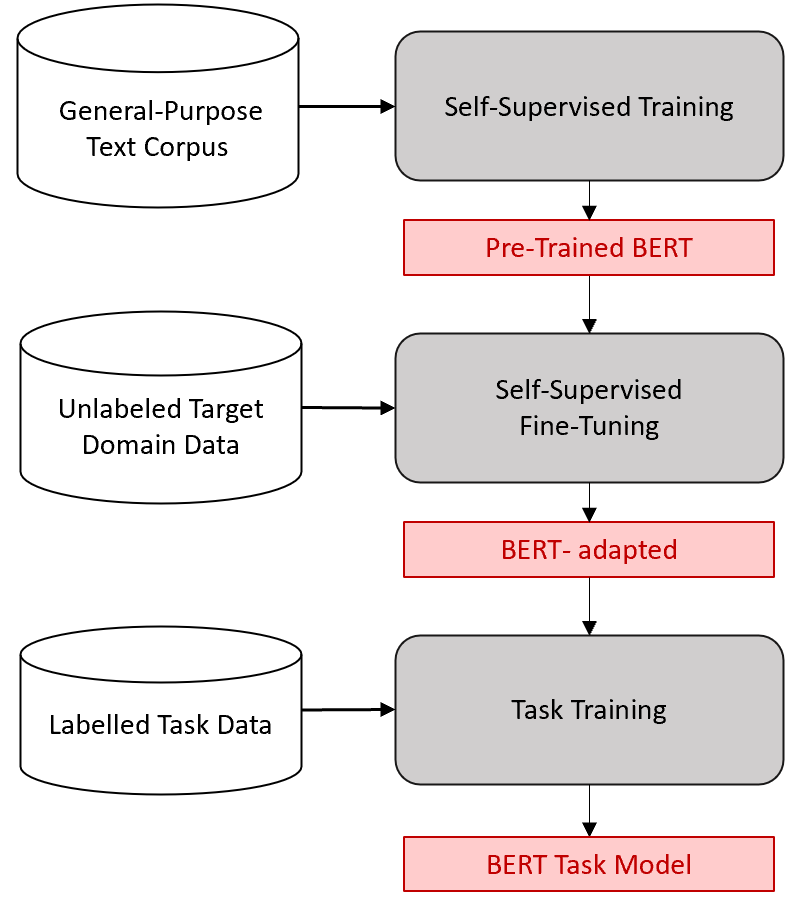}
    \caption{Our training process.}
    \label{fig:process}
\end{figure}

Recently, contextualized word embeddings such as ELMo \cite{elmo} and BERT \cite{bert}, trained on large data, have demonstrated remarkable performance gains across many NLP tasks. In our experiments, we use BERT as the base model that serves as the input into our duplicate question detection task model. To improve low-resource DQD, both within- and cross-domain, we address learning a domain-adapted DQD task model within a two stage approach starting from pretrained BERT. This process is depicted in \figref{process}.

In the first stage, because BERT is pretrained on general purpose text from Wikipedia and books which are largely different from user generated posts in the \stackexchange{} domains, we explore unsupervised domain adaptation of the base BERT model to a new domain. The adaptation of BERT to scientific \cite{sciBERT}, biomedical \cite{bioBERT}, and historical English \cite{adaptaBERT2019} domains has previously shown promising performance. Following a similar approach, we adapt the existing pretrained BERT model using self-supervised objectives of masked language modeling and next sentence prediction on unlabeled data from StackExchange. In the second stage, the BERT-adapted model on the target domain is then finetuned on the DQD task data and objectives to train the BERT task model.

We experiment on four domains related to computer systems and two other non-related domains, and show the effectiveness of our proposed approach.
Our main findings are as follows: 
(i) We show that BERT helps the task of DQD in \stackexchange{} domains and outperforms the previous LSTM-based baseline.
(ii) Our unsupervised adaptation of BERT on unlabeled domain data improves
the results substantially, especially in low-resource settings where
less labeled training data is available.
(iii) We show that adapting BERT on even a small amount of unlabeled data from target domains is very effective.
(iv) We demonstrate that unsupervised adaptation of BERT
on a large number of diverse \stackexchange{} domains further improves performance for a variety of target domains.

% alrochet: we should probably reference hal daume's Direct Approach for the BERT baseline

\section{Unsupervised Domain Adaptation of Contextual Embeddings: Background}
Contextual embeddings are pretrained on large, topically-diverse text to learn generic representations useful for various downstream tasks. However, the effectiveness of these models decreases as the mismatch between the pretraining material and the task domains increases.
To alleviate this issue, ULMFiT \cite{howard2018} trains LSTM models on generic unsupervised data, 
but then fine-tunes them on domain data before training task-specific models from them.
When large amounts of unlabeled domain data is available, this domain adaptation step provides significant improvements even with only small amounts ($<$1$k$) of labeled training examples. 
We propose a similar process but adopt BERT \cite{bert} and apply it to our DQD task.
We find that adaptation, even on little unlabeled data
from  the target domain, is effective.

Adapting BERT to target domains is also studied in several recent works. 
BioBERT \cite{bioBERT} shows significant improvement for a suit of biomedical tasks 
by fine-tuning BERT on large biomedical data.
SciBERT \cite{sciBERT} does similarly but for scientific domains. 
AdaptaBERT \cite{adaptaBERT2019} fine-tunes BERT to unlabeled historical English text in a domain adaptation scenario where training data comes from contemporary corpora. 
AdaptaBERT significantly improves BERT for POS tagging in this setting.
Similar to this line of related work, we also fine-tune BERT on domain specific data
but for the task of DQD in community forums. 
We complement this line of research with novel findings.
% Our scenarios go beyond just fine-tuning the pretrained contextual embeddings like BERT on a new task -- which is well studied in literature.
For example, our scenario of unsupervised domain adaptation using small unlabeled domain data 
% (see \figref{trainingsize-unsup})
has not been addressed in any prior work in BERT (or other contextual models), to the best of our knowledge.
This is a critical result given the importance of contextual models, because it shows
for low-resource domains there is a big room for improvements even if they have small unlabeled data.
Also for cross-domain, we are first to apply BERT to a semantic task (i.e., DQD).
The other work \cite{adaptaBERT2019} addresses POS tagging -- a low-level task in NLP.

\section{BERT for Duplicate Question Detection in Community Forums}
\seclabel{model}
The task is to find whether two questions are duplicate or not,
based on their semantic similarity. 
To compute the similarity, we follow a similar setup as the BERT's sequence pair classification experiments in \newcite{bert}.
First, we learn a representation for the pair of questions.
Then, a linear binary classifier with cross entropy loss is applied to the representation.

A question includes a title and a body. To obtain a representation for each pair of questions $(q_1, q_2)$, we compute two vectors: one for the titles and one for the bodies. Each vector is generated from the last-layer's [CLS] token's embedding, 
starting from an input of the form $[\mbox{CLS}] S_{q_1} [\mbox{SEP}] S_{q_2} [\mbox{SEP}]$, 
where $S_{q}$ takes the value of the title or the body for question $q$. [\mbox{CLS}] and [\mbox{SEP}] are special tokens in BERT. Similar to \cite{shah2018adversarial}, the two vectors are then summed to produce a single representation for the pair of questions. 

%First, the two questions are concatenated to form the input to BERT.
%Each question includes a ``title'' and a ``body''.

%We add the pooled vector of each (i.e., last-layer's [CLS] token's embedding) and sum them to get one representation for the pair of questions.

%% TJ: what is CLS?

\subsection{Low-Resource Settings}
We aim to improve generalization of DQD in low-resource settings.
In \stackexchange{}, many domains have limited data, e.g., 
%TODO
``Android'' domain has 48,490 posts with around 2,000 duplicate annotations. % TJ: Fill in X and Y
Using domains in \stackexchange{} such as ``AskUbuntu'' or ``Android'',
we propose procedures to improve within- and cross-domain generalization for DQD.
Our primary setup has small or no labeled within-domain data, limited unlabeled within-domain data, and large out-of-domain unlabeled data.
When there is no within-domain labeled data, we use labeled data from other domains to learn the task.
%This setup is common when applying DQD as an application.

\subsection{Unsupervised BERT Domain Adaptation}
Previous work in DQD \cite{shah2018adversarial} has addressed domain adaptation when 
no within-domain training data is available. 
They train randomly initialized BiLSTMs, with pretrained word embeddings on \stackexchange{} as input, and apply adversarial objectives for domain adaptation.
% alrochet: I'm not sure i follow the sentence below. Maybe this could work "In this work, we make use of unlabeled data to learn a representation using self-supervised objectives. This is used as an adaptation to the general domain of question answering as well as a way to do transfer learning across specific \stackexchange domains."
%
We instead focus on transfer learning to adapt pre-trained models to new target domains. %YY: Is it better now?
Our models are based on BERT, which has proven effective for many different NLP tasks. BERT is pre-trained on unlabeled data using two self-supervised objectives: masked language modeling (MLM) and next sentence prediction (NSP). 
The corpora BERT is trained on are: 2.5B tokens of Wikipedia and .8B tokens of BookCorpus \cite{Zhu_2015_ICCV}.
% Many target domains have very different 
% lexical and syntactic properties compared to these corpora which 
% are mostly written and edited by professionals. 
%The recent work shows promising improvements over BERT by 
%either fine-tuning or training BERT on domain specific text.
% We test two levels of adaptation when pretraining BERT. 
% First adaptation is to the StackExchange in general. 
% Second adaptation is to the specific domain of StackExchange.
% Also, we test our models in domain adaptation and within-domain training.

Our domains are taken from \stackexchange{}, which includes various topics, such as sports, travel, food, programming, etc.
The posts are written by diverse internet users,
with variations in their vocabulary and syntax compared to BERT's pretraining corpora written and edited mostly by professionals. 
Therefore, we adapt BERT by fine-tuning it on \stackexchange{} posts using
the same self-supervised objectives of MLM and NSP. We refer to this model as \textbf{BERT-adapted} as opposed to the original \textbf{BERT} model. %% TJ- Do we want to refer to BERT-base instead of just BERT?

\section{Experiments}

\subsection{Baseline Experiments}

\begin{table}[]
\small
\resizebox{\columnwidth}{!}{\begin{tabular}{lllll}
\hline
Dataset    & Unsupervised & Train & Dev   & Test  \\ \hline
AskUbuntu  & 305,769      & 9,106 & 1,000 & 1,000 \\
SuperUser  & 390,378      & 9,106 & 1,000 & 1,000 \\
Apple      & 93,399       & 2,000 & 1,000 & 1,000 \\
Android    & 48,490       & -     & 1,000 & 1,000 \\ \hline
% 20-domains & 1,162,487    & -     & -     & -     \\
% 33-domains & 1,531,797    & -     & -     & -     \\ \hline
\end{tabular}}
\caption{Datasets statistics. The Unsupervised column indicates the number of questions. Train, Test and Dev columns specify the number of positive duplicates.}
\label{tab:datasets}
\end{table}

We use the DQD \textbf{datasets} of \newcite{shah2018adversarial} as well as their training and testing protocol. 
Our target domains are (AskUbuntu, Android, Apple, SuperUser).
The data contains pairs of questions.
The positive examples are taken from the duplicate marks in \stackexchange{}. Unlabeled examples are extracted from the \stackexchange{} dumps.\footnote{https://archive.org/details/stackexchange}
We append the body to the title for each question and make that a contiguous paragraph for BERT self-supervised adaptation.\footnote{We use the Pytorch implementation \url{https://github.com/huggingface/pytorch-pretrained-BERT}}
Some statistics are shown in \tabref{datasets}.
For unsupervised adaptation of BERT, we also use unlabeled data sets from additional \stackexchange{} domains. Along with the datasets for the specific target domains, we craft two additional data sets from 20 and from 33 different stackexchange{} domains (including the four target domains). We list the selected 20 and 33 domains with total data amounts in the Appendix.

Since the annotations are incomplete, \newcite{shah2018adversarial} propose to use AUC as the \textbf{metric} for DQD performance since it is more robust against false negatives.
They report the normalized AUC@0.05, which is the area under the curve of the true positive rate as function of the false positive rate ($fpr$), from $fpr = 0$ to $fpr = 0.05$.
We follow the same protocol and use AUC@0.05 metric. 

\begin{table*}[t]
\small
\centering
\begin{tabular}{r|ll|c|c|cccccc}
                   &  &                     &          & & \multicolumn{6}{c}{\textbf{BERT-adapted}}                \\
                   
ln& \textbf{Source}              & \textbf{Target}              & \textbf{Baseline} & \textbf{BERT}    & Target  & 20d   & 33d & 20d-noTarget & 20d-noNSP & 20d-frozen  \\
\hline
1 & Apple               & Android    & .764     & .826     & .883  & .910 & {.919}  & .889    & .866 & .771\\
2 & AskUbuntu           & Android    & .790     & .810     & .907  & .883 &  {.909 } & .849    & .863 & .830\\
3 & SuperUser           & Android    & .790     & .849     & .908  & .907 & {.914} & .886    & .891 & .805\\
\hline
4& AskUbuntu            & Apple          & .855     & .857      & .927  & .931 &  {.942} & .916   & .916 &.889\\
5 & SuperUser           & Apple      	 & .861     & .881      & .939  & .943 &  {.954} & .931   & .929 & .887\\
\rowcolor{Gray}
6 & Apple               & Apple       	 & .976     & .982      & .989  & .988 &  .991 &  .991   & .987 & .870\\
\hline
7 & Apple               & AskUbuntu  	& .756     & .683      & .864  & .872 &  {.897} & .812   & .833 & .767\\
8 & SuperUser           & AskUbuntu     & .796    & .779       & .870  & .874 &  {.891} & .826   & .835 & .812\\
\rowcolor{Gray}
9 & AskUbuntu           & AskUbuntu     & .858    & .899       & .923  & .920 &  {.942} & .921   & .924 & .845\\
\hline
10 & Apple               & SuperUser    & .873    & .925       & .965  & .968 &  {.973} & .959   & .958 & .916\\
11 & AskUbuntu           & SuperUser    & .911    & .932       & .964  & .966 &  {.975} & .957   & .955 & .937\\
\rowcolor{Gray}
12 & SuperUser           & SuperUser    & .930    & .958       & .967  & .974 &  {.977} & .968   & .967 & .937\\
\hline
13 & \multicolumn{2}{l|}{Avg (Source $\neq$ Target)}     
                                             & .822    & .838       & .914  & .917 &  {.930} &  .892 & .897 & .846\\
\rowcolor{Gray}
14 & \multicolumn{2}{l|}{Avg (Source $=$ Target)}     
                                             & .921    & .946       & .960  & .961 &  {.970}  & .960 &.959 & .884\\
\hline
15 & \multicolumn{2}{c|}{Overall Avg}        & .847    & .865       & .926 & .928  &  {.940}  & .909 & .912 & .855
\end{tabular}
\caption{AUC@0.05 for the baseline \cite{shah2018adversarial}, 
BERT, and BERT-adapted models. 
BERT-adapted models are shown by the domain(s) they adapted on.
% 20d and 30d are listed in \tabref{domains}.
Results of BERT-adapted on 20d-noTarget and Target in each row correspond to one of the four target models, depending on the row's target domain.
The rows with white (gray) background belong to cross(within)-domain experiments.}
\tablabel{big}
\end{table*}

\begin{figure*}[h]
\begin{subfigure}{.5\textwidth}
  \centering
  \includegraphics[width=180pt,height=120pt]{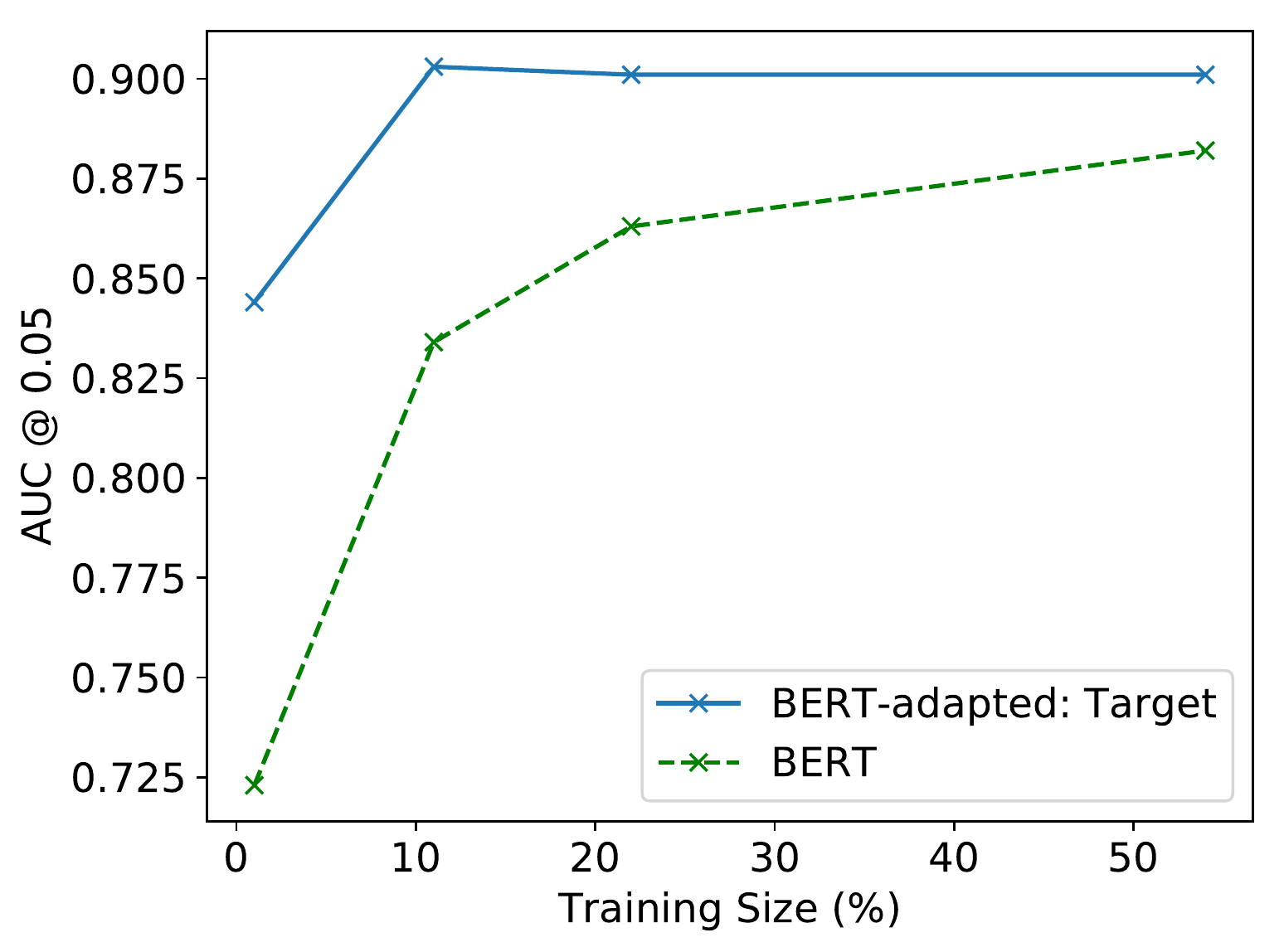}
  \caption{AskUbuntu$\rightarrow$AskUbuntu}
  \figlabel{askubuntu}
\end{subfigure}
~
\begin{subfigure}{.5\textwidth}
  \centering
   \includegraphics[width=180pt,height=120pt]{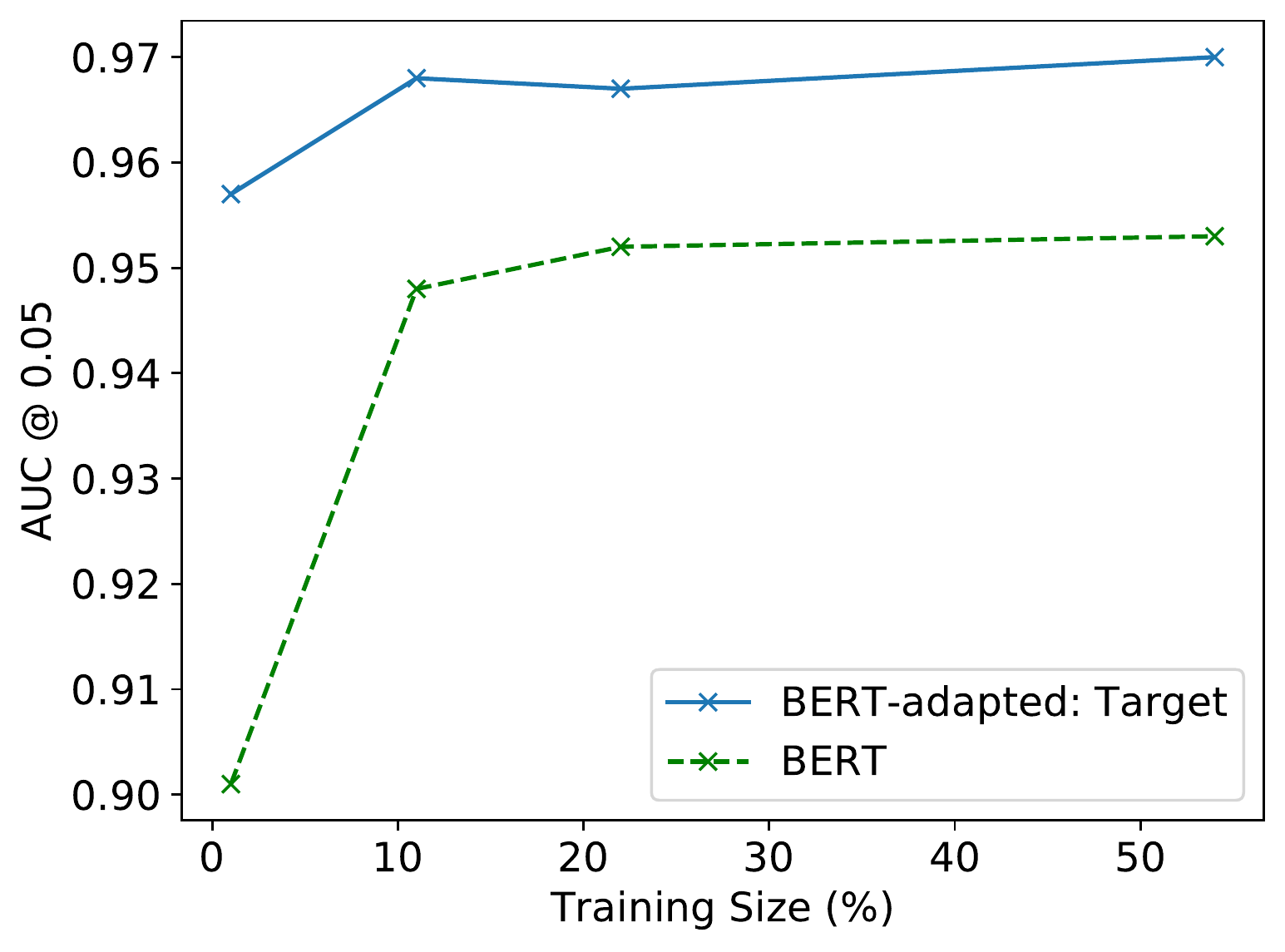}
  \caption{SuperUser$\rightarrow$SuperUser}
  \figlabel{superuser}
\end{subfigure}
 \caption{AUC@0.05 vs training set size for AskUbuntu$\rightarrow$AskUbuntu (a)
 and SuperUser$\rightarrow$SuperUser (b).
  After 50\%, all graphs remain flat.}
\figlabel{trainingsize-sup}
\end{figure*}

\begin{figure}[h]
\centering
  \includegraphics[width=180pt,height=120pt]{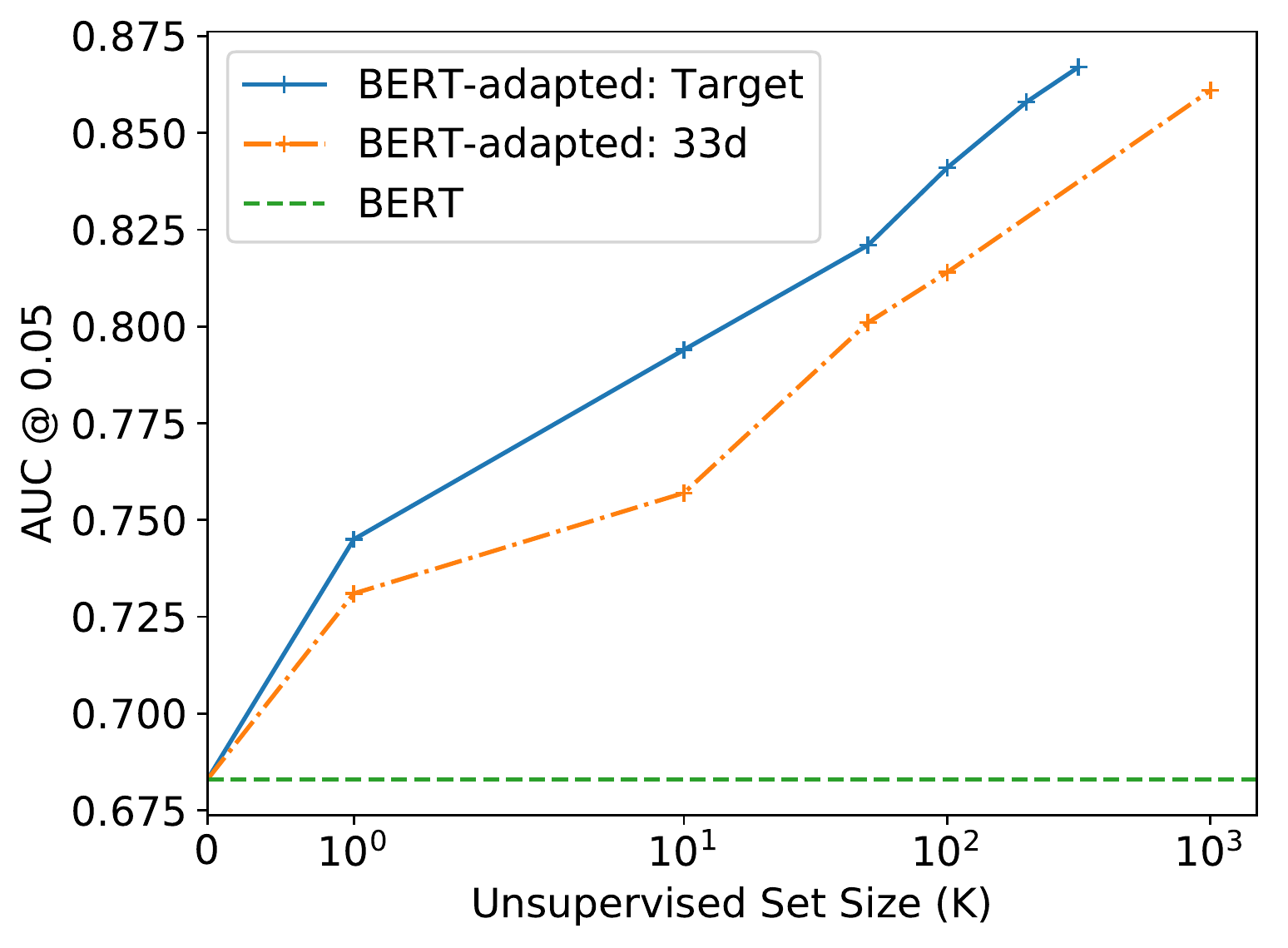}
  \caption{AUC@0.05 vs unsupervised data size for the Apple$\rightarrow$AskUbuntu cross-domain scenario.}
  \label{fig:trainingsize-unsup}
\end{figure}
%TODO YY: update the curves' labels.

\textbf{Results.}
\tabref{big} shows the main results for 12 different combinations (3 within-
and 9 cross-domain) of training a DQD model using a source domain and testing on a target domain, with early stopping on the dev set of
the target domain, similar to \newcite{shah2018adversarial}.
We also add three average rows for cross-domain (line 13), within-domain (line 14), and overall (line 15).
We use \newcite{shah2018adversarial}'s model as our \textbf{baseline}.
For the scenarios not covered in the baseline's paper, we run
their published code.\footnote{To confirm, we were able to replicate the paper's results
using their published code.}
For BERT, we use the BASE-CASED pretrained model with a fixed set of hyperparameters for task fine-tuning.

%\tabref{big} shows our main experiment results. We examine 12 different combinations of training a DQD model using source domain and testing on a target domain. Of these 12 combinations, three are within-domain scenarios and nine are cross domain scenarios. We also add 3 rows for the cross-domain, within-domain, and overall average results.
%For each source/target condition multiple initial models are explored.
%We use \newcite{shah2018adversarial}'s model as our \textbf{baseline}.
%For the setups not covered in the baseline's paper, we run
%their published code.\footnote{To confirm, we were able regenerate the paper's results
%using their published code.}
%For BERT, we use the BASE-CASED pretrained model and use the default hyperparameters for fine-tuning it. The remaining columns cover different BERT-adapted variants.

As \tabref{big} demonstrates, BERT outperforms the baseline on average. 
However, it is worse in two cases (lines 7 \& 8).
BERT-adapted on the target domain is better than BERT and the baseline in all rows.
It substantially improves BERT for cross-domain (.838 to .914).
This indicates that BERT-adapted learns better representations
for the target-specific terms compared to BERT by its unsupervised training on unlabeled target data.\footnote{We also check the effect of seeing target's test examples in the unsupervised fine-tuning. 
We analyze BERT-adapted on Apple$\rightarrow$AskUbuntu, and the results show that for the same size of data, whether or not the unlabeled test questions were included had no effect on the DQD performance.}
%Note that for each of the four tagets, a separate BERT-adapted models is trained, and used for the corresponding rows with that target.
BERT-adapted by 20 domains (20d) does not degrade the performance 
of BERT-target on average.
Note that this model is trained only once and used in all rows.
We further see improvements by adding more domains: BERT-adapted on 33 domains
outperforms all models consistently. 

Excluding the row's target domain from 20d hurts the performance 
for the cross-domain cases. (see 20d-noTarget column). 
However, it still outperforms BERT, emphasizing that related unlabeled data is also beneficial for adaptation.
This is also clear in the effectiveness of 33 domains which spans more related data.

Another investigation is on the contribution of next sentence prediction (NSP)
objective in the performance of BERT-adapted.
Our 20d-noNSP results show that removing NSP in adaptation only decreases the overall average from .928 to .912, still significantly higher than BERT (.865). 

We also analyze what happens if we freeze the BERT-adapted parameters when
we train the task (20d-frozen in \tabref{big}).
Only the final-layer DQD classifier parameters have to be learned in this case.
%In \tabref{analysis} we observe that, 
We experiment on the 12 scenarios, the frozen BERT-adapted results are significantly worse than their fine-tuned counterparts, but, interestingly, they are almost as good as the fine-tuned BERT models (i.e., models where BERT parameters are also updated) which are not adapted to target domains.	

\subsection{Limited Labeled Data Scenario}

In \figref{trainingsize-sup}, we vary the DQD training size and report the performance of the AskUbuntu$\rightarrow$AskUbuntu (\figref{askubuntu}) and SuperUser$\rightarrow$SuperUser  (\figref{superuser}) scenarios,
performing supervised DQD training starting form both the BERT and target-specific BERT-adapted models. The notation $Domain_A$$\rightarrow$$Domain_B$ indicates that we train on $Domain_A$ and evaluate on $Domain_B$.
For both datasets, BERT-adapted models tuned on only 1\% of labeled
data achieve better results than the BERT models tuned on 10\%. For AskUbuntu the BERT-adapted model trained for DQD on only 1\% of the labeled data even beats the BERT model trained for DQD on the full labeled data set. 
% BERT-adapted tuned on only 1\% of labeled DQD training data achieves .825, 
% significantly bigger than .725. 
These results demonstrate that unsupervised adaptation of BERT to a domain significantly reduces the need for annotated data for supervised training of the DQD task. As we increase the training size, the gap between BERT and 
BERT-adapted shrinks. This confirms our assumption that adapting BERT is especially effective for sparse annotated data scenarios.

\subsection{Limited Unlabeled Data Scenario}

In \figref{trainingsize-unsup}, we show how the DQD performance in the Apple$\rightarrow$AskUbuntu scenario improves when increasing the size of randomly sampled unlabeled data (number of questions) from
only AskUbuntu or 33 different domains. 

BERT-adapted models are improved as more unlabeled data is added, but even small amounts (1$k$) of unlabeled data is helpful.
Another interesting observation is that, even though sampling unsupervised examples from the target domain appears to be superior, the ability to sample more data from the combined set of 33 domains closes the gap (See \figref{trainingsize-unsup}) and eventually achieves superior performance (see  \tabref{big}).

\begin{table*}[t]
\small
\centering
\begin{tabular}{l|cccccc}
          & Quora & AskUbuntu & Apple & Android & SuperUser & Academia \\
          \hline
% Sprint    & 0.044 & 1.000  &           &       &         &           &          \\
Quora     & 1.000 &     -     &    -  &   -     &  -        &  -        \\
AskUbuntu & 0.123 & 1.000     &     - &   -     &  -        &   -       \\
Apple     & 0.154 & 0.169     & 1.000 &   -     &   -       &    -      \\
Android   & 0.134 & 0.128     & 0.240 & 1.000   &   -       &    -      \\
SuperUser & 0.131 & 0.221     & 0.174 & 0.118   & 1.000     &     -     \\
Academia  & 0.157 & 0.085     & 0.168 & 0.173   & 0.083     & 1.000   
\end{tabular}
\caption{Vocab Jaccard Index between datasets' question titles.}
\tablabel{domainSimilarity}
\end{table*}

\begin{table}
\small
\centering
\begin{tabular}{l|cc|cc}
          & \multicolumn{2}{c}{Unigram} & \multicolumn{2}{c}{Bigram} \\
          & Positive     & Negative     & Positive     & Negative    \\
\hline
AskUbuntu & 0.160        & 0.030        & 0.049        & 0.003\\       
Apple     & 0.170        & 0.027        & 0.053        & 0.003\\       
Android   & 0.157        & 0.031        & 0.044        & 0.003\\     
SuperUser & 0.188        & 0.027        & 0.064        & 0.003\\   
Academia  & 0.125        & 0.040        & 0.028        & 0.002 \\ 
Quora     & \textbf{0.468}        & \textbf{0.302}        & \textbf{0.248}        & \textbf{0.152}       
\end{tabular}
\caption{Vocab Jaccard Index between question title pairs in each dataset. 
This shows how Quora's annotation differ from StackExchange datasets
in terms of lexical similarity between question pairs.}
\tablabel{pairSimilarity}
\end{table}

\begin{table}[h]
\small
    \centering
    \begin{tabular}{ll | c| c c }
         & & & \multicolumn{2}{c}{\textbf{BERT-adapted}} \\
         \textbf{Source} & \textbf{Target} & \textbf{BERT} & \textbf{Target} & \textbf{Target-frozen} \\
         \hline \hline
         Academia & AskUbuntu & .601 & .854 & .758\\
         Quora    & AskUbuntu & .515 & .609 & .670\\
         \hline
         SuperUser & AskUbuntu & .779 & .870 & .821 \\
         AskUbuntu & AskUbuntu & .899 & .923 & .845 \\
        %  Sprint   & AskUbuntu & .562 & .749\\
    \end{tabular}
    \caption{Comparing AUC(0.05) DQD results of AskUbuntu when source training data comes from topically different
    domain (Academia and Quora) and when the non-duplicate distribution is quite different (Quora)
    with a related domain (SuperUser) and the target domain itself (AskUbuntu).}
    \tablabel{quoraAsk}
\end{table}

\subsection{Large Domain Variation Scenario}

To examine the effectiveness of adaptation when the domain difference between source and target domains is large, 
we examine performance on AskUbuntu with two additional datasets for training: Quora and Academia.
Academia is similarly built from SE as datasets in \newcite{shah2018adversarial}.
Quora is taken from \newcite{shah2018adversarial}. The annotation of Quora comes from the released
Quora question pairs dataset.\footnote{{https://data.quora.com/First-Quora-Dataset-Release-Question-
Pairs}}

In \tabref{domainSimilarity}, we show the lexical similarity between questions in different datasets. 
We also measure the similarity of pairs for positive (duplicate) vs negative (non-duplicate) examples, shown in \tabref{pairSimilarity}.
Focusing on AskUbuntu, we see that Academia and Quora both have low similarity with AskUbuntu. 
That is more clear for Academia where the vocab Jaccard Index is only 0.085.
In addition to lexical variation with AskUbuntu, Quora has another significant difference: its negative pairs were deliberately selected to have high lexical overlap, while for AskUbuntu (and other SE datasets here) the negative examples are chosen randomly. 
It results in the vocabulary Jaccard Index between duplicate/non-duplicate questions being  much higher in Quora (0.468/0.302)
compared to Academia (0.125/0.040) and other SE domains.
This gives Quora's labeled annotations different distributional characteristic than those in SE. 
Basically, the labeling function of Quora is different from the labeling function used in the SE datasets.

In \tabref{quoraAsk}, we show the results of domain adaptation from Quora and Academia to AskUbuntu, 
for BERT and BERT-adapted on AskUbuntu (fine-tuned or frozen BERT) as the target domain.
To understand how these results are compared when more related domains are used for training, 
we also include the results of SuperUser (as the best performing source domain for AskUbuntu) 
and AskUbuntu itself.

As it is shown, the base BERT works poorly when either Academia or Quora is the source dataset for DQD training.
Results of Academia improve substantially when BERT is adapted on AskUbuntu and fine-tuned for DQD on Academia training data. In fact, performance from training on Academia is only slightly worse than training on the more topically-similar domain of SuperUser (0.854 vs. 0.870). Freezing BERT-adapted parameters before training on the DQD on Academia decreases the performance, similar to SuperUser or AskUbuntu results. For Quora, however, BERT-adapted results are still very low and interestingly freezing BERT parameters performs better. 
We believe this is primarily caused by the difference in the labeling function used in Quora, i.e., the model trained on Quora is learning a different query similarity definition which does not generalize well to SE.

In summary, our results show that for domain-adaptation DQD: (i) if the labeling function is similar, the behaviour for topically-different domains is similar to cases 
where domains are topically similar; (ii) when training data comes from a dataset with a different labeling function, it is better to freeze the BERT parameters and only update the classification layer.

\section{Discussion and Conclusion}
In this paper, we introduced a new process to improve the task of low-resource DQD.
In general, our process includes two main steps: 
(i) domain adaptation of BERT on unlabeled data using its self-supervised objectives and
(ii) use the adapted BERT from step (i) and fine-tune it on a DQD dataset using the supervised objective (cross-entropy). 
The main focus of the paper is on (i), and how it affects the task results.
Through extensive evaluation with BERT, we showed that this adaptation improves the
generalization of DQD models greatly when there is no or limited training data available
for the target domain.
Two main scenarios of within- and cross-domain were addressed for evaluation
on five domains of StackExchange and also Quora.
% The main element of our process is to adapt contextual embeddings, like BERT, by self-supervised learning on the unlabeled data from target domains.

Our results suggest that: 
\begin{itemize}
\item A combination of unsupervised training on a target domain with supervised training on a different source domain is an effective strategy for the DQD task.
\item Significantly less supervised DQD training data is needed if we first adapt BERT with unsupervised training to data from the target domain.
\item Unsupervised adaptation of BERT on even small amount of target data yields  better models.
\item If the annotation function of the source material is different than the target material, then domain adaptation faces issues which cannot be fixed by unsupervised adaptation.
% In this case, fixing the BERT parameters during 
% task training shows better performance compared to standard fine-tuning.
\end{itemize}

As future work, we plan to apply our process to other similar scenarios; we believe our approach is generally effective, especially for low-resource applications.
% alrochet: Maybe cite adverserial stuff we want/have tried
As another direction, applying other techniques like adversarial domain adaptation to BERT while learning the task seems as a complementary component to our method. Our initial attempt on this by using simple approaches did not make any improvement, but more investigation is needed which we leave for future work.

\bibliography{refs}
\bibliographystyle{IEEEbib}

\section*{Appendix: List of domains}
\begin{table}[h]
\caption*{List of the 20 and 33 domains used in our unsupervised BERT adaptation experiments.}\centering
\begin{tabular}{l|p{2.1in}l}
\textbf{20 domains} \\ (1,162,487 posts)
    & academia
android
apple
askubuntu
aviation
bitcoin
boardgames
christianity
cooking
cs
gaming
hinduism
judaism
linguistics
mechanics
meta.superuser
philosophy
politics
superuser
workplace
 \\
\textbf{33 domains} \\ (1,531,797 posts) & \textbf{20 domains} + anime
astronomy
bicycles
biology
buddhism
chemistry
cogsci
crypto
islam
meta.stackexchange
skeptics
sports
unix
 
\end{tabular}
    
    \label{tab:domains}
\end{table}

\end{document}